\documentclass{article}

\usepackage[final]{neurips_2025}

\usepackage[utf8]{inputenc} % allow utf-8 input
\usepackage[T1]{fontenc}    % use 8-bit T1 fonts
\usepackage{hyperref}       % hyperlinks
\usepackage{url}            % simple URL typesetting
\usepackage{booktabs}       % professional-quality tables
\usepackage{wrapfig} 
\usepackage{amsfonts}       % blackboard math symbols
\usepackage{nicefrac}       % compact symbols for 1/2, etc.
\usepackage{microtype}      % microtypography
\usepackage{xcolor}         % colors
\usepackage{graphicx}
\usepackage{multirow}

\usepackage{bm}
\usepackage{amsmath}
\usepackage{hyperref}
\usepackage{caption}

\title{REArtGS: Reconstructing and Generating Articulated Objects via 3D Gaussian Splatting with Geometric and Motion Constraints}

\author{
Di Wu$^{1,2}$ , Liu Liu $^{3}$\thanks{Corresponding author} ,
Zhou Linli$^{1}$, Anran Huang$^{3}$, Liangtu Song$^{1}$, Qiaojun Yu$^{4}$, \\
\textbf{Qi Wu}$^{4, 5}$\thanks{Project leader}  , \textbf{Cewu Lu}$^{4}$ \\
1 \textit{Hefei Institutes of Physical Science Chinese Academy of Sciences}
 \\
2 \textit{University of Science and Technology of China} \\
3 \textit{Hefei University of Technology} \\
4 \textit{Shanghai Jiao Tong University}\\
5 \textit{ByteDance} \\
Email: \texttt{wdcs@mail.ustc.edu.cn, liuliu@hfut.edu.cn} 
}

\begin{document}

\maketitle

\begin{abstract}
 Articulated objects, as prevalent entities in human life, their 3D representations play crucial roles across various applications. However, achieving both high-fidelity textured surface reconstruction and dynamic generation for articulated objects remains challenging for existing methods. In this paper, we present REArtGS, a novel framework that introduces additional geometric and motion constraints to 3D Gaussian primitives, enabling realistic surface reconstruction and generation for articulated objects. Specifically, given multi-view RGB images of arbitrary two states of articulated objects,  we first introduce an unbiased Signed Distance Field (SDF) guidance to regularize Gaussian opacity fields, enhancing geometry constraints and improving surface reconstruction quality. Then we establish deformable fields for 3D Gaussians constrained by the kinematic structures of articulated objects, achieving unsupervised generation of surface meshes in unseen states. Extensive experiments on both synthetic and real datasets demonstrate our approach achieves high-quality textured surface reconstruction for given states, and enables high-fidelity surface generation for unseen states. Project site: \href{https://sites.google.com/view/reartgs/home}{\texttt{https://sites.google.com/view/reartgs/home}}.
\end{abstract}

\section{Introduction}
\label{sec:intro}
Articulated objects are ubiquitous in our daily lives. Modeling articulated objects, i.e. mesh reconstruction and generation, holds significant importance in many fields of computer vision and robotics fields including virtual and augmented reality~\cite{liu2022toward,xue2021omad}, object manipulation~\cite{yu2024gamma,wang2024rpmart} and human-object interaction~\cite{yang2022oakink,li2024semgrasp}. Currently, the surface and shape reconstruction for articulated objects is a non-trivial task due to the challenges: firstly, articulated objects exhibit complicated and diverse geometric structures with a wide range of scales. Secondly, articulated objects possess varied kinematic structures, and the static surface meshes generated from vanilla 3D reconstruction methods fail to meet practical interaction requirements. 
Under this circumstance,  PARIS~\cite{liu2023paris} attempts to use multi-view images from two states for articulated object dynamically reconstruction with neural implicit radiance fields. Nevertheless, PARIS lacks geometric constraints, leading to shape-radiance ambiguity and additional errors for motion analysis. 

\begin{wrapfigure}[15]{R}{0.5\textwidth}
%\begin{figure}[t]
\centering
\includegraphics[width=1.0\linewidth]{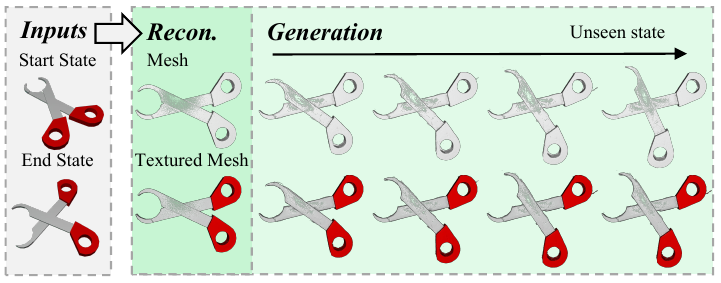}
\caption{Given multi-view RGB images of articulated objects from two arbitrary states, our REArtGS enables high-quality textured surface mesh reconstruction and generation for unseen states.}
\label{fig:teaser}
%\vspace{-0.5cm}
%\end{figure}
\end{wrapfigure}
% \begin{wrapfigure}{r}{0.5\textwidth} % 右侧对齐，占半栏宽度
%   \centering
%   \includegraphics[width=0.9\linewidth]{figs/teaser.pdf}
%   \caption{Given multi-view RGB images of articulated objects from two arbitrary states, our REArtGS enables high-quality textured surface mesh reconstruction and generation for unseen states.}
%   \label{fig:teaser}
% \end{wrapfigure}

In recent years, 3D Gaussian Splatting (3DGS)~\cite{kerbl20233d} achieves realistic and real-time novel view synthesis through explicit representation with 3D Gaussian primitives. %Benefiting from 3DGS, some researchers start to conduct dynamic reconstruction~\cite{li2024dgns,duan20244d,liu2024dynamic}.
Some subsequent works have expanded 3DGS to surface reconstruction~\cite{yu2024gaussian, huang20242d, chen2024pgsr} and dynamic reconstruction ~\cite{yang2024deformable, liu2024dynamic, duan20244d}. However, existing 3DGS-based surface reconstruction methods typically suffer from insufficient geometric constraints~\cite{yu2024gaussian} or impose constraints by restricting the shapes of Gaussian primitives ~\cite{huang20242d, guedon2024sugar}, leading to noisy surface reconstruction results. On the other hand, current dynamic reconstruction methods via 3DGS tend to input motion time and spatial positions  into neural networks to obtain the deformed positions of Gaussian primitives~\cite{yang2024deformable, duan20244d}. Consequently, these methods require continuous supervision throughout the entire motion, which limits their ability to generate surface meshes in unseen states. In general, introducing 3DGS as a ready-to-use technique into articulated object surface reconstruction and generation is still not feasible and remains a challenging task.

In this paper, to address the aforementioned issues and make full advantage of 3DGS, we propose \textbf{REArtGS}, a novel approach that \textbf{R}econstructs and g\textbf{E}nerates high-quality texture-rich mesh surfaces for \textbf{Art}iculated objects via 3D \textbf{G}aussian \textbf{S}platting, only taking multi-view RGB images from two arbitrary states. Specifically,
unlike most 3DGS methods based on the opacity field, which may lead to noisy surface extraction due to the non-strict linearity of the opacity field, we first introduce the Signed Distance Function (SDF) field to facilitate geometry learning. Then we propose a geometry constraint which encourages the SDF values of the Gaussian primitives approach zero when their opacity values reach the maximum , achieving unbiased guidance of the SDF field over the opacity field.  In this way, we can take advantage of the strong linearity of the SDF field to explicitly establish a connection between Gaussian opacity field and the scene surface, which significantly enhances geometric learning and  reduces artifacts.

Subsequently, we employ the optimized Gaussian primitives as an accurate geometry initialization for dynamic surface generation. Due to the absence of intermediate states, we leverage the kinematic structures of articulated objects to model time-continuous Gaussian deformable fields in an unsupervised fashion, and constrain the deformable fields using learnable motion parameters. Concretely, we propose a heuristic method for unsupervised part segmentation and formulate the deformable fields of dynamic parts via the motion parameters.  Our REArtGS is evaluated on PartNet-Mobility~\cite{xiang2020sapien} and AKB-48~\cite{liu2022akb} object repositories, ranging from synthetic to real-world data. Extensive experiments demonstrate that our REArtGS outperforms existing state-of-the-art methods in articulated object surface reconstruction and generation tasks, as shown in Fig.~\ref{fig:teaser}.

In summary, our main contributions can be summarized as follows: (1) We propose REArtGS, a novel framework introducing 3DGS to conduct high-quality textured surface reconstruction and time-continuous generation for articulated objects, only using multi-view images from two arbitrary states. (2) Our REArtGS exploits an unbiased SDF guidance for 3D Gaussian primitives to enhance geometric constraints for improving reconstruction quality, and also establishes the deformable fields constrained by kinematic structures of articulated objects to generate unseen states in an unsupervised manner. (3) We incorporate REArtGS into various scenes ranging from synthetic to the real-world data across many different articulation categories. The extensive experimental results demonstrate that our approach significantly outperforms SOTAs in both mesh reconstruction and generation tasks.

\vspace{-1.0em}
\section{Related Work}
\vspace{-0.5em}
\subsection{Articulated Object Shape Reconstruction}

% DeepSDF, A-SDF, Ditto, CARTO, 
% Object surface reconstruction is a well-established problem for understanding the full geometric shape of objects. Some works introduce to encode continuous functions that model the objects using Signed Distance~\cite{park2019deepsdf}, radiance~\cite{mildenhall2021nerf, rakotosaona2024nerfmeshing, tang2023delicate} and occupancy~\cite{mescheder2019occupancy}. Going beyond static scenes with rigid objects, A-SDF~\cite{mu2021sdf} extends the DeepSDF to learn a latent code for modeling an articulated object by both shape and motion. Following up on A-SDF, CenterSnap~\cite{irshad2022centersnap} designs a point cloud decoder for shape reconstruction, and Shapo~\cite{irshad2022shapo} fully utilizes shape prior for latent space learning. To generalize these works into the real world, AKBNet~\cite{liu2022akb} integrates the shape reconstruction module into the object knowledge modeling along with part segmentation and pose estimation. To achieve both geometry and motion analysis in a single forward, Ditto~\cite{jiang2022ditto} and REACTO~\cite{song2024reacto} propose to generate shapes at unseen states from pair observations. Currently, PARIS~\cite{liu2023paris} alleviate the 3D data requirement limitation and success to reconstruct shape surface with only RGB images. But it still suffer from several challenges such as lacking constraints modeling and failing in texture generation. Therefore, our REArtGS aims to address these issues, and proposes geometric and motion constrained 3DGS for this task.
Object surface reconstruction is a well-established problem for understanding the full geometric shape of objects. Some works introduce to encode continuous functions that model the objects using Signed Distance~\cite{park2019deepsdf, mu2021sdf}, radiance~\cite{mildenhall2021nerf, rakotosaona2024nerfmeshing, tang2023delicate} and occupancy~\cite{mescheder2019occupancy}. To achieve both geometry and motion analysis in a single forward, Ditto~\cite{jiang2022ditto} and REACTO~\cite{song2024reacto} propose to generate shapes at unseen states from pair observations. Specifically, PARIS~\cite{liu2023paris} alleviates the 3D data requirement limitation and succeeds in reconstructing shape surface with only RGB images. Most recently, ArticulatedGS~\cite{guo2025articulatedgs} and ArtGS~\cite{liu2025building}  introduce 3DGS to achieve both the reconstruction and motion estimation of articulated objects.  However, these methods  still suffer from insufficient geometry constraints, which may lead to noisy surface mesh outputs. Therefore, our REArtGS aims to address these issues, and proposes geometric and motion constrained 3DGS for this task.
\subsection{Surface Reconstruction with 3DGS}
3D Gaussain Splatting has become increasingly popular technique for surface reconstruction in recent years~\cite{kerbl20233d, li2024loopgaussian, yu2024mip, qin2024langsplat}. GOF~\cite{yu2024gaussian} establishes opacity fields of 3D Gaussians using ray-tracing-based rendering, and extracts the surface meshes by the opacity level set. 2DGS~\cite{huang20242d} and PGSR~\cite{chen2024pgsr} seek to transform 3D Gaussians into 2D flat representation, obtaining accurate normal distribution. Although improving the surface reconstruction quality, these methods still lack more reasonable geometry constraints. Several researchers attempt to integrate SDF representation with 3D Gaussians~\cite{yu2024gsdf, chen2023neusg, xiang2024gaussianroom, dai2024high}, but these approaches commonly use SDF to regularize normals and guide the pruning of 3D Gaussians, without substantial optimization of Gaussian opacity fields.
Meanwhile, some works conduct dynamic reconstruction using 3DGS. Deformable 3DGS proposes a deformation field to reconstruct dynamic scenes with 3D Gaussian primitives. 4DGS~\cite{duan20244d} leverages a spatial-temporal structure encoder and a multi-head Gaussian deformation decoder to derive the deformed 3D Gaussians at a given time. However, these methods rely on complete supervision of the motion process, limiting their capacity for generating unseen states. The similar issues also occur in DGMesh~\cite{liu2024dynamic} and REACTO~\cite{song2024reacto}. We provide more analysis and comparison of the related works in  Appendix.

%\section{Problem Statement and Notations}
% 仿照PARIS的3.Problem Statement来写，注意notation的定义要明确
%Our REArtGS aims to solve the problem of articulated object surface reconstruction and generation. The problem is defined as follows: taking RGB image collections from two arbitrary states $s=0$, $s=1$ of the object, reconstructing high-fidelity textured meshes and generating high-quality textured meshes for any unseen states between $s \in [0,1]$.
%For textured mesh reconstruction, we propose an unbiased SDF-guided Gaussian splatting approach to reconstruct dense point clouds at the initial motion state, providing an accurate geometric prior for dynamic reconstruction. Second, we perform dynamic reconstruction in a self-supervised fashion by learning the motion parameters of articulated objects. The details of our approach are elaborated below.
\begin{figure}[tbh]
    \centering
    \vspace{-0.5em}
    \includegraphics[width=0.9\linewidth]{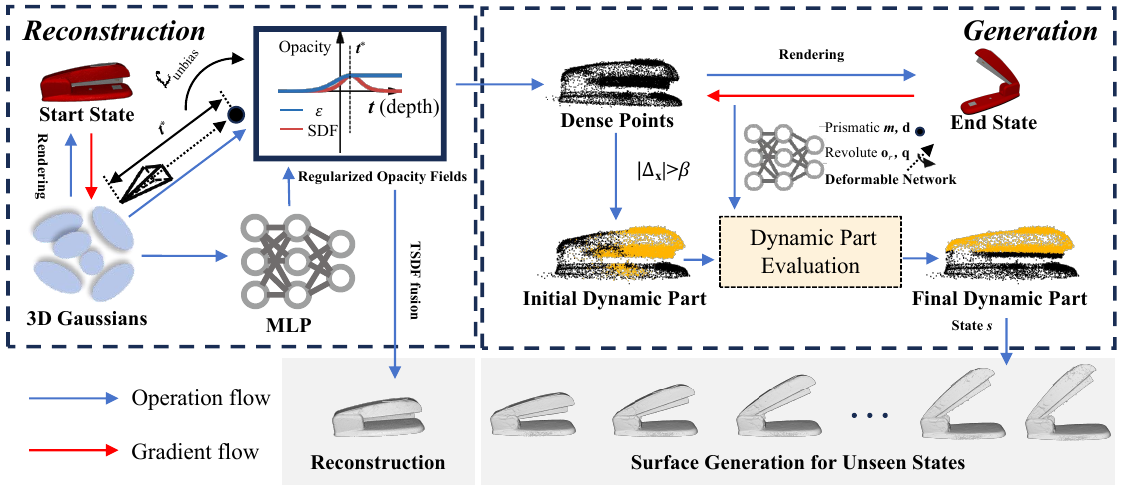}
    % \vspace{-1.8em}

    \caption{The overall pipeline of REArtGS. We introduce additional geometric and motion constraints for 3D Gaussian primitives, achieving high-quality surface mesh reconstruction and time-continuous generation, with only multi-view images from arbitrary two states.}
    \label{fig:method}
    \vspace{-0.5cm}
\end{figure}

\section{Method}
%核心方法：SDF-gudied regularized opacity field, 基于运动学参数的高斯形变场 deformable field
The overall framework of our REArtGS is illustrated in Fig.~\ref{fig:method}. Taking multi-view RGB images from two arbitrary states $s=0$, $s=1$ of articulated objects, we aim to achieve high-quality textured mesh reconstruction and generation at any unseen states between $s \in [0,1]$.

We first introduce SDF representation and propose an unbiased SDF regularization to enhance the geometry constraints of 3D Gaussian primitives. In this manner, we improve the reconstruction quality and yield dense point clouds at state $s=0$, providing an accurate geometric prior for subsequent dynamic reconstruction. Then we establish time-continuous deformable fields for 3D Gaussian primitives constrained by the kinematic structures of articulated objects. Given any state $s$, we can derive the deformed position $x_{s}$ of Gaussian primitives through the deformation fields in an unsupervised fashion. The details of our approach are elaborated below.
 %First, we propose an unbiased SDF-guided Gaussian splatting approach to reconstruct dense point clouds at the initial motion state, providing an accurate geometric base for dynamic reconstruction. Second, we perform dynamic reconstruction in a self-supervised fashion by learning the motion parameters of articulated objects. The details of our approach are elaborated below. 

%todo:修改字体
% \begin{figure}[tbh]
%     \centering
%     % \vspace{-1.2em}
%     \includegraphics[width=0.8\linewidth]{figs/unbiased_SDF1.pdf}
%     % \vspace{-1.8em}
    
%     \caption{The illustration of unbiased SDF regularization. It can be observed that when $t$ approaches $t^{*}$, the absolute value of SDF $|\mathcal{S}|$ converges to zero with the unbiased SDF regularization.} %$\bm{\hat{\sigma}}$  the unbiased SDF regularization significantly reduces the spatial deviation between the maximum point of  value and the zero point of SDF . 
%        \label{fig:unbiased SDF}
%        \vspace{-0.5cm}
% \end{figure} 
\begin{wrapfigure}[15]{R}{0.5\textwidth} % 右侧对齐，占半栏宽度
  \centering
  \vspace{-0.5cm}
  \includegraphics[width=\linewidth]{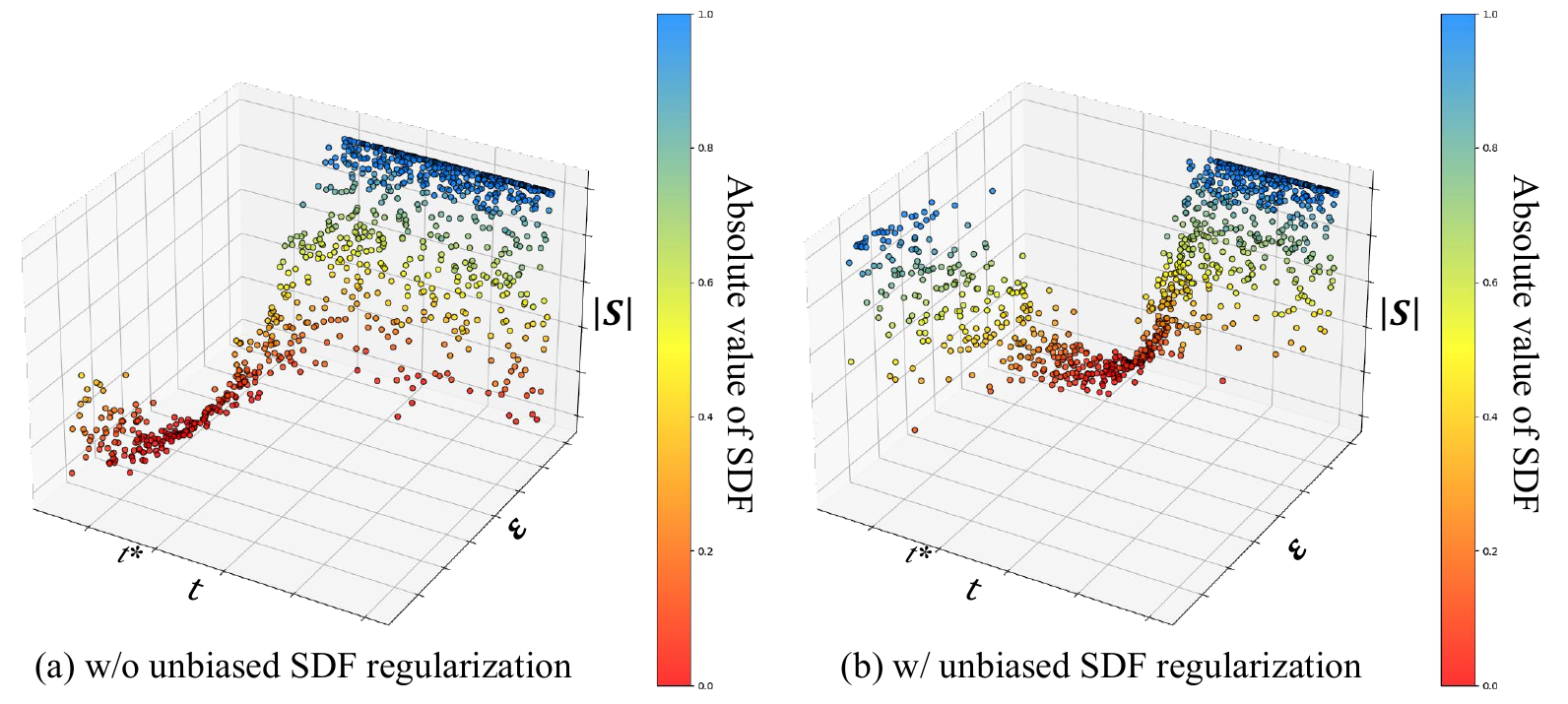}
  \caption{The illustration of unbiased SDF regularization. It can be observed that when $t$ approaches $t^{*}$, the absolute value of SDF $|\mathcal{S}|$ converges to zero with the unbiased SDF regularization.} 
  \label{fig:unbiased SDF}
\end{wrapfigure}

\subsection{Reconstruction with Unbiased SDF Guidance}
\label{sec:stage one}
%In the first stage, we conduct static surface reconstruction for the initial motion state based on the ray-marching-based 3DGS, to provide a high-quality geometry base for next dynamic reconstruction. 
To better preserve the geometric features of Gaussian primitives, we use the ray-tracing-based 3DGS following GOF, which evaluates rendering contribution of Gaussian primitives directly without the 3D-to-2D projection step. However, the Gaussian primitives still suffer from insufficient geometric constraints. Concretely, the opacity fields lack an explicit link with scene surface, which may lead to noisy outputs in surface reconstruction of articulated objects. 
%to perform scene reconstruction, which leads to the following challenges: 1. The opacity has no explicit relationship with the scene surface and lacks geometric constraints for Gaussian primitives. 2. The normal distributions of Gaussian primitives are difficult to estimate accurately, since their shapes are not limited. 
%The learnable opacity coefficient $\alpha$ lacks a more reasonable initialization, and the dimension of its search space increases exponentially with the complexity of the model.

To address the limitation, we first introduce SDF representation to guide the geometry learning of Gaussian primitives. We utilize a 
Multi-layer Perceptrons (MLP) with 8 hidden layers to learn the SDF values for spatial position inputs, and the scene surface can be represented by the zero-set:
\begin{equation}
\mathcal{S}=\left\{\mathbf{x} \in \mathbb{R}^{3} \mid f(\mathbf{x})=0\right\}
\end{equation}

Following Neus~\cite{wang2021neus}, we can derive the opacity $\bm{\hat{\sigma}}$ for the center $\mathbf{x}_{i}$ of Gaussian $G_{i}$ from learned SDF values:
\begin{equation}
    \label{eq:sdf2opacity}
	\bm{\hat{\sigma}}_{i}=\text{max} \left(\frac{\Phi\left(f\left(\mathbf{x}_{i}\right)\right)-\Phi\left(f\left(\mathbf{x}_{i+1}\right)\right)}{\Phi  \left(f\left(\mathbf{x}_{i}\right)\right)}, 0\right)
\end{equation}
where $\Phi$ denotes a Sigmoid function. Nevertheless, the SDF-opacity conversion is not directly applicable to the $\alpha$ blending rendering. This is because each 3D Gaussian primitive models a local density distribution, while Eq.~\ref{eq:sdf2opacity} only maps the global positions to their opacity values and neglects the local properties of Gaussian primitives. Therefore, we define the opacity of Gaussian primitive combining the rendering contribution $\bm{\varepsilon}$ proposed in GOF as:
\begin{equation}
\label{eq:opacity}
\begin{aligned}
\bm{\sigma_{i}}&= \bm{\hat{\sigma}}_{i}\bm{\varepsilon}\left(G_{i}\right) \\
               &= \bm{\hat{\sigma}}_{i}\cdot\text{max}(e^{- \frac{1}{2} {\mathbf{x}_{L}}  ^{T}\mathbf{x}_{L}}) \\
               &= \bm{\hat{\sigma}}_{i}\cdot\text{max}\left(e^{-\frac{1}{2}\left(\mathbf{r}_{L}^{T} \mathbf{r}_{L} t^{2}+2 \mathbf{o}_{L}^{T} \mathbf{r}_{L} t+\mathbf{o}_{L}^{T} \mathbf{o}_{L}\right)}\right)
\end{aligned}
\end{equation}
where $\mathbf{o}_{L}$, $\mathbf{r}_{L}$ are the camera center and incident ray direction represented in the local coordinate system of Gaussian $G_{i}$, which can be acquired by local scaling matrix and rotation matrix of $G_{i}$.
%We can acquire them by the local scaling matrix $\mathbf{S} \in \mathbb{R} ^{3\times3} $, and rotation matrix $\mathbf{R}\in \mathbb{R} ^{3\times3} $ of Gaussian $G_{i}$ following GOF:
%\begin{equation}
%\begin{aligned}
%\mathbf{o}_{L} & =\mathbf{S}_{i}^{-1} %\mathbf{R}_{i}\left(\mathbf{o}-\bm{\mu}_{i}\right) \\
%\mathbf{r}_{L} & =\mathbf{S}_{i}^{-1} \mathbf{R}_{i} \mathbf{r},
%\end{aligned}
%\end{equation}
The maximum value of $\bm{\varepsilon}\left(G_{i}\right)$ occurs at ray depth $t^{*}$, expressed as follows:
\begin{equation}
t^{*} = \frac{\mathbf{o}_{L}^{T}\mathbf{r}_{L}}{\mathbf{r}_{L}^{T}\mathbf{r}_{L}}
\end{equation}
The rendering equation based on $\bm{\sigma}$ is represented as:
\begin{equation}
\label{eq:rendering}
\mathbf{C}=\sum_{i =1}^{N} \mathbf{c}_{i} \bm{\sigma_{i}} \prod_{j=1}^{i-1}\left(1-\bm{\sigma_{i}}\right)
\end{equation}
where $ N $ is the number of Gaussian primitives involved in $\alpha$-blending and $ \mathbf{c}_{i} $ is the color modeled with spherical harmonics. To leverage SDF to constrain the geometric learning of Gaussian primitives, we first introduce a bell-shaped function $\Phi_{k}$ to modulate the transformation from SDF to opacity and replace the original Sigmoid activation function $\Phi$ in Eq.~\ref{eq:sdf2opacity}, formulated as:
\begin{equation}
\label{eq:activation}
\Phi_{k}(f(\mathbf{x}))=\frac{e^{k \cdot f(\mathbf{x})}}{\left(1+e^{k \cdot f(\mathbf{x})}\right)^{2}}
\end{equation}
where $k$ is a learnable parameter that adjusts the function shape and we initially set it to 0.1. %Intuitively, the maximum value of $\Phi_{k}(f(\mathbf{x}))$ is achieved when the SDF value of Gaussian primitive approaches zero, that is, the Gaussian primitive closer to the surface has a higher opacity value. 
Intuitively, the Gaussian primitive closer to the surface has a higher opacity value. 

However,  since $\bm{\hat{\sigma}}$ is determined solely by the Gaussian's center,  a non-alignment still exists between the maximum points of $\bm{\hat{\sigma}}$ and $\bm{\varepsilon}$, which can be formulated as:
\begin{equation}
    t_{\text{bias}}=\left \| \underset{t}{\text{argmax}}(\bm{\hat{\sigma}})-\underset{t}{\text{argmax}}(\bm{\varepsilon)}  \right \| 
\end{equation}

To mitigate this bias, we propose an unbiased regularization for the SDF value at depth $t^{*}$ as following:
\begin{equation}
    \mathcal{L}_{\text{unbias}}= \left \| f(\mathbf{o}+t^{*} \mathbf{r} ) \right \|_{2}^{2}
\end{equation}
where $\mathbf{o}$, $\mathbf{r}$ are the camera center and incident ray direction. Through the unbiased regularization, we encourage that when the rendering contribution of a Gaussian primitive reaches its maximum value, the corresponding spatial position is close to the scene surface, as shown in Fig.~\ref{fig:unbiased SDF}.  In this manner, the SDF representation is able to regularize the opacity fields without bias and facilitate more reasonable distribution of Gaussian primitives over scene surface. We provide more detailed elaboration and proof in the Appendix.
%Note that although 3DGSR [] uses a similar function as Eq. \ref{eq:activation}, it ignores the local features of Gaussian primitives, potentially resulting in unstable reconstruction quality.

\subsection{Mesh Generation with Motion Constraints}
\label{sec:stage two}
Due to the explicit representation of Gaussian primitives, we can  acquire accurate dense point clouds from reconstruction, serving as initial geometry for mesh generation. Then we establish time-continuous deformable fields for 3D Gaussians to infer the deformed position $\mathbf{x}_{s}$ at state $s$. However, it is challenging to optimize the deformable fields for Gaussian primitives only with the supervision of two states. Accordingly, we exploit the kinematic structures of articulated objects to constrain the Gaussian deformation field. We first assume that the articulated motion only includes rotation and translation following PARIS. Then we employ a heuristic method to efficiently predict the joint type through movement trend of points, which is elaborated in the Appendix. For rotation joints, our learning objective is to determine their rotation axis and rotation angle. Specifically, we take the pivot point $\mathbf{o}_{r}\in \mathbb{R}^{3}$ of the rotation axis and a normalized quaternion $\mathbf{q} \in \mathbb{R}^{4}$ as learnable parameters, where the quaternion can be decoupled into the rotation axis $\mathbf{a}\in \mathbb{R}^{3}$ and rotation angle $\theta $. For prismatic joints, our learning objective is to determine their translation directions $\mathbf{d}$ and translation distances $m$. Similarly, we take the unit vector $\mathbf{d} \in \mathbb{R}^{3}$ representing the translation direction and the translation distance $m$ as learnable parameters.

Subsequently, we formulate the deformable fields through a canonical state. Given the rotation motion $SO(3) \in \mathbb{R} ^{3\times3} $, we define its canonical state as $s^{*}=0.5$ and the rotation angle of state $s$ can be expresses as:
\begin{equation}
\theta_{s} =  \frac{\left ( s^{*} -s \right ) }{s^{*}} \theta
\end{equation}
where $s\in\left [ 0,1 \right ] $ and $\theta\in\left [ - \pi /2, \pi /2 \right ] $. The angle-bounded parameterization is able to prevent the singularity in exponential coordinates when $\left \| \theta  \right \| > \pi$. Using Rodrigues' rotation formula~\cite{pina2011rotations}, the deformed position $\mathbf{x}_{s}$  can be derived as following:
\begin{equation}
  \mathbf{x}_{s} =  \left (  \mathbf{I} + \sin(\theta_{s} )\mathbf{K} + (1 - \cos(\theta_{s} ))\mathbf{K}^{2}\right ) \left ( \mathbf{x}-\mathbf{o}_{r}\right ) +\mathbf{o}_{r}
\end{equation}
where $\mathbf{K}$ is a skew-symmetric matrix formed by the rotation rotation axis $\mathbf{a}$ and $\mathbf{I}\in \mathbb{R} ^{3\times3}$ is a unit matrix. Meanwhile, we align the local quaternions of 3D Gaussian primitives after rotation through quaternion multiplication.

For the prismatic motion $m\cdot\mathbf{d} \in \mathbb{R} ^{3}$, we take advantage of the vector space structure of Euclidean geometry by setting $s^{*}=0$. The deformed position $\mathbf{x}_{s}$ can be naturally acquired by linear interpolation, formulated as:
\begin{equation}
\mathbf{x}_{s} = \mathbf{x} + s\cdot m \cdot \mathbf{d}
\end{equation}
unlike rotation motion, this parameterization avoids singularities due to the flat Riemannian structure of $\mathbb{R} ^{3}$.
 
However, the deformable fields should be applied exclusively to movable Gaussian primitives. Consequently, the Gaussian primitives need to be segmented into dynamic and static parts. To address this challenge, we propose an unsupervised method for dynamic part segmentation. We first only input the multi-view images at the end state during warm-up training iterations, and then perform an initialization segmentation of the dynamic part. The segmentation criterion for the initial dynamic components is represented as: $ \left |\Delta_{\mathbf{x}}\right |>\beta $, where $ \left |\Delta_{\mathbf{x}}\right | $ is the L1 distance of spatial position variation during the warm-up training, and $\beta$ is its average value. Furthermore, the spatial transformation of the dynamic part should align with the learnable motion parameters. Hence the dynamic part among the Gaussian primitives is re-evaluated at certain intervals of iterations, where the criterion for revolute or prismatic joints is represented as:
\begin{equation}
\begin{aligned}
\left | \hat{\theta} - \theta \right | &<\frac{ \varphi_{\theta}}{K} \\
\left | \hat{m} - m \right | &<\frac{ \varphi_{m}}{K}
\end{aligned}
\end{equation}
where $\hat{\theta}$, $\hat{m}$  are the rotation angle around axis $\mathbf{a}$  and the translation distance respectively, $\varphi$ is the tolerance threshold which decreases as the number of iterations $K$ increases.

Moreover, our method can be conveniently transferred to multi-part articulated objects, through sequentially learning the segmentation masks and motions of each individual part. Please refer to our Appendix for more details.
\subsection{Optimization and Textured Mesh Extraction}
\begin{figure}[t]
    \centering
    % \vspace{-1.2em}
    \includegraphics[width=0.9\linewidth]{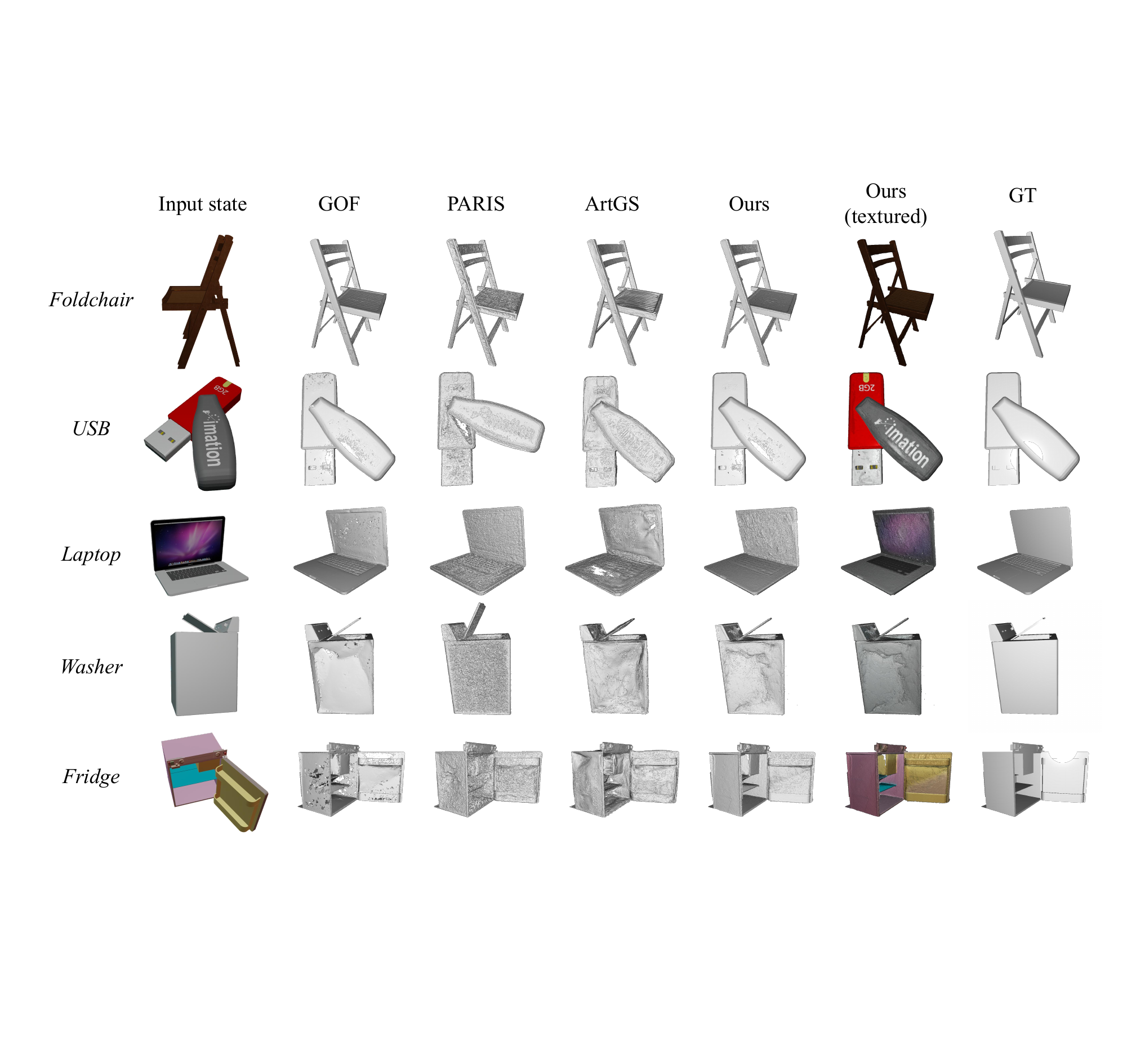}

    \caption{The qualitative result of surface reconstruction on PartNet-Mobility dataset. We show both textured and non-textured meshes for the best comparison.}
       \label{fig:mesh_reconstruction}
\vspace{-2.0em}
\end{figure}

% todo:没有介绍SDF正则法线的优势
Since the shapes of Gaussian primitives are not limited, the normal distributions of Gaussian primitives are difficult to estimate accurately. However,  Gaussian's normals should be parallel to the gradients of SDF in an ideal case. We consequently introduce additional regularization to regularize the normals of Gaussian primitives by SDF representation, expressed as follows:
\begin{equation}
	\label{euqa:normal}
	\mathcal{L}_{\text{normal}} = \frac{1}{N} \sum_{i}^{N} (1-\frac{\left|\mathbf{n}_{i} \cdot \nabla f(\mathbf{x}_{i}) \right|}{\left\|\mathbf{n}_{i}\right\| \cdot\left\|f(\mathbf{x}_{i})\right\|})
\end{equation}
where $\mathbf{n}_{i}$ represents the normal of Gaussian $G_{i}$, which is calculated following GOF:
\begin{equation}
	\label{euqa:normal}
	\mathbf{n}_{i} = -R_{i}^{T}S_{i}^{-1}\mathbf{r}_{L}
\end{equation}
where $R$ is the rotation matrix encoded by Gaussian's quaternion and $S$ is the scaling matrix. We also employ the Eikonal regularization term~\cite{gropp2020implicit} to encourage SDF gradient direction perpendicular to the surface, represented as:
\begin{equation}
	\label{euqa:eik}
	\mathcal{L}_{\text{eik}}=\frac{1}{N} \sum_{i=1}^{N}\left(\left\|\nabla f(\mathbf{x}_{i}))\right\|_{2}-1\right)^{2}
\end{equation}

The overall training objective $\mathcal{L}$ is formulated as follows:
\begin{equation}
\mathcal{L} = \mathcal{L}_{c} + \lambda_{1}\mathcal{L}_{\text{unbias}} + \lambda_{2}\mathcal{L}_{\text{normal}} + \lambda_{3}\mathcal{L}_{\text{eik}} +\lambda_{3}\mathcal{L}_{d}
\end{equation}
where $\lambda$ is the weight of regularization and $\mathcal{L}_{d}$ is the depth distortion loss following \cite{huang20242d}. $\mathcal{L}_{c}$ is defined as:
\begin{equation}
\mathcal{L}_{c} = \lambda\mathcal{L}_{\text{D-SSIM}}+(1-\lambda)\mathcal{L}_{1}
\end{equation}
where $\mathcal{L}_{\text{D-SSIM}}$ is proposed in \cite{kerbl20233d} and $\mathcal{L}_{1}$ is the L1 norm of the pixel loss.

Once the optimization converges, we adopt the TSDF fusion algorithm\cite{newcombe2011kinectfusion} to extract the textured mesh from Gaussian primitives. We utilize the $\alpha$-blending rendering pipeline to obtain depth, opacity and RGB renderings for training views. Then we integrate these images into a voxel block grid (VBG) and extract a triangle mesh from the VBG. The textures of surface meshes can be efficiently obtained  leveraging the spherical harmonics and opacity of Gaussian primitives. Please refer to our Appendix for a more detailed illustration.

% \begin{figure*}[t!]
%     \centering
%     % \vspace{-1.2em}
%     \includegraphics[width=\linewidth]{figs/generation.pdf}
%     % \vspace{-1.8em}

%     \caption{\textbf{The qualitative result of surface generation on PartNet-Mobility dataset. We obtain surface textured meshes at end states from canonical states through the deformation fields and conduct a visual comparison of the surface mesh quality against existing state-of-the-art methods.} }
%        \label{fig:generation}
% \end{figure*} 

%\uppercase\expandafter{\romannumeral1}
\begin{table*}[tbh]
\centering
\caption{Quantitative results for the surface reconstruction quality on PartNet-Mobility dataset. We bold the best results and underline the second best results. $*$ means we implement it without depth supervision for fair comparison.}
\resizebox{0.95\linewidth}{!}{
\begin{tabular}{cc|ccccccccccc}
\toprule
Metrics   &Method      & Stapler & USB &  Scissor & Fridge & Foldchair & Washer & Blade & Laptop & Oven & Storage & Mean \\
\hline

\multirow{6}{*}{CD(ws) $\downarrow$} 
& A-SDF~\cite{mu2021sdf} &14.19 &7.14 &10.61 &13.71 &40.85 &12.50 &3.31 &2.11 &21.37 &22.57 &14.84 \\
& Ditto~\cite{jiang2022ditto} & 2.38 &2.09 & 1.70 &2.16 &6.80 &\textbf{7.29} &42.04 &\underline{0.31} &\textbf{2.51} &\textbf{3.91} &7.19 \\
& PARIS~\cite{liu2023paris} & \textbf{0.96}  & \underline{1.80}  & \underline{0.30}  &2.68  &0.42  &18.31 &\textbf{0.46} &\textbf{0.25} &\underline{6.07} &\underline{8.12} &\underline{3.94} \\
& GOF~\cite{yu2024gaussian} &\underline{1.51} &8.09 &0.36 &\underline{1.51} &0.53 & 17.35&0.89 &  0.84& 19.81&10.04 &6.09\\
& ArtGS$^{*}$~\cite{liu2025building} &2.77 &1.36 &0.75 &2.01 &\underline{0.41} &20.59 &\underline{0.63} &0.99 &9.01 &9.00 &4.75 \\
& REArtGS (Ours) &3.47 & \textbf{0.75}&\textbf{0.29} &\textbf{1.50} & \textbf{0.40}&\underline{12.20} &0.72 &0.53 &8.89 &8.27 & \textbf{3.79}\\
\hline

\multirow{6}{*}{CD(rs) $\downarrow$}  
& A-SDF~\cite{park2019deepsdf} &\underline{2.140} &2.478 &\textbf{1.805} &1.801 &2.080 &1.590 &5.630 &2.418 &2.751 & 4.077&2.677 \\
& Ditto~\cite{jiang2022ditto} & 2.874 &3.049 & 2.926 &1.550 &0.925 &\underline{1.428} &5.792 &1.184 &1.296 &2.837 & 2.386\\
& PARIS~\cite{liu2023paris} & \textbf{2.077}&3.812 &2.807 &\textbf{0.370} &1.209 &3.869 & 2.855&1.025 &2.873 & 2.903&2.380\\
& GOF~\cite{yu2024gaussian}        &  2.510 &1.736 &2.304 &2.230 &2.787 &5.725 & \underline{2.111} &\underline{0.961} &3.080 &2.812 &2.626\\
& ArtGS$^{*}$~\cite{liu2025building} &2.949 &\underline{1.704} &2.438 &0.725 & \underline{0.615} & 2.874 & \textbf{2.031} &1.182 &\underline{1.148} &\textbf{1.157} &\underline{1.682}\\
& REArtGS (Ours) &2.186 & \textbf{1.433}& \underline{2.291}&\underline{0.475} & \textbf{0.018}& \textbf{1.204}&2.596 &\textbf{0.038} & \textbf{0.784}& \underline{1.330}&\textbf{1.236}\\
\hline
\multirow{6}{*}{F1 $\uparrow$}
& A-SDF~\cite{park2019deepsdf} &0.041 &0.035 &0.094 &0.019 & 0.053&0.046 &0.224 & 0.001& 0.010&0.015 &0.054 \\
& Ditto~\cite{jiang2022ditto} &0.197 &0.181 &0.275 &0.114 & 0.352 &0.059 &0.107 &0.366 &0.052 &0.030 &0.173 \\
& PARIS~\cite{liu2023paris} & 0.240 &0.151 &0.343 &0.091 &0.429 &0.024 & 0.396 &\textbf{0.533} &0.031 & 0.033&0.227\\
& GOF~\cite{yu2024gaussian}   &0.217 & 0.189& \textbf{0.614} &\textbf{0.173} & \underline{0.480} &\textbf{0.087} & 0.386 &0.303 & 0.049 & 0.036&\underline{0.253}\\
& ArtGS$^{*}$~\cite{liu2025building} &\underline{0.251} &\underline{0.209} &0.438 &0.118 &0.447 &0.053 &\underline{0.408} &0.294 &\underline{0.056} &\underline{0.042} &0.232 \\
& REArtGS (Ours)  &\textbf{0.256} & \textbf{0.307}& \underline{0.598} &\underline{0.165} & \textbf{0.502}& \underline{0.069} & \textbf{0.488} & \underline{0.419} &\textbf{0.065} & \textbf{0.066}&\textbf{0.294}\\
\hline
\multirow{6}{*}{EMD $\downarrow$} 
& A-SDF~\cite{park2019deepsdf} &\textbf{0.755} & 1.113 & 0.952 &0.945 &1.020 & 0.982 &1.763 & 1.039& 1.174&1.258 &1.100 \\
& Ditto~\cite{jiang2022ditto} &1.724 &1.308 &1.212 &0.619 &0.935 &\underline{0.841} &1.996 &0.417 &\underline{0.852} &0.970 &1.087 \\
& PARIS~\cite{liu2023paris} &1.197 &1.381 & \textbf{0.637} & \textbf{0.426} & 0.778 &1.660 & 1.823&0.706 & 1.200 & 1.046& 1.085\\
& GOF~\cite{yu2024gaussian} & 1.643&1.815 &1.075 &1.055 &1.181 &1.692 & 1.237 & \underline{0.394} &1.241 & 0.959 &1.229\\
& ArtGS$^{*}$~\cite{liu2025building} &1.921 &\textbf{0.677} &\underline{0.934} &0.637 &\underline{0.506} &1.154 &\underline{1.131} &0.502 &1.006 &\underline{0.787} &\underline{0.926}\\
& REArtGS (Ours) &\underline{1.060} & \underline{0.847} &1.078 & \underline{0.485} &\textbf{0.097} &\textbf{0.777} &\textbf{1.112} &\textbf{0.111} &\textbf{0.627} & \textbf{0.755}& \textbf{0.695}\\
\bottomrule
\end{tabular}}
\label{tab:exp_reconstruction}
\vspace{-0.3cm}
\end{table*}

\begin{figure*}[tbh]
\centering
%\vspace{-0.5em}
\includegraphics[width=0.9\linewidth]{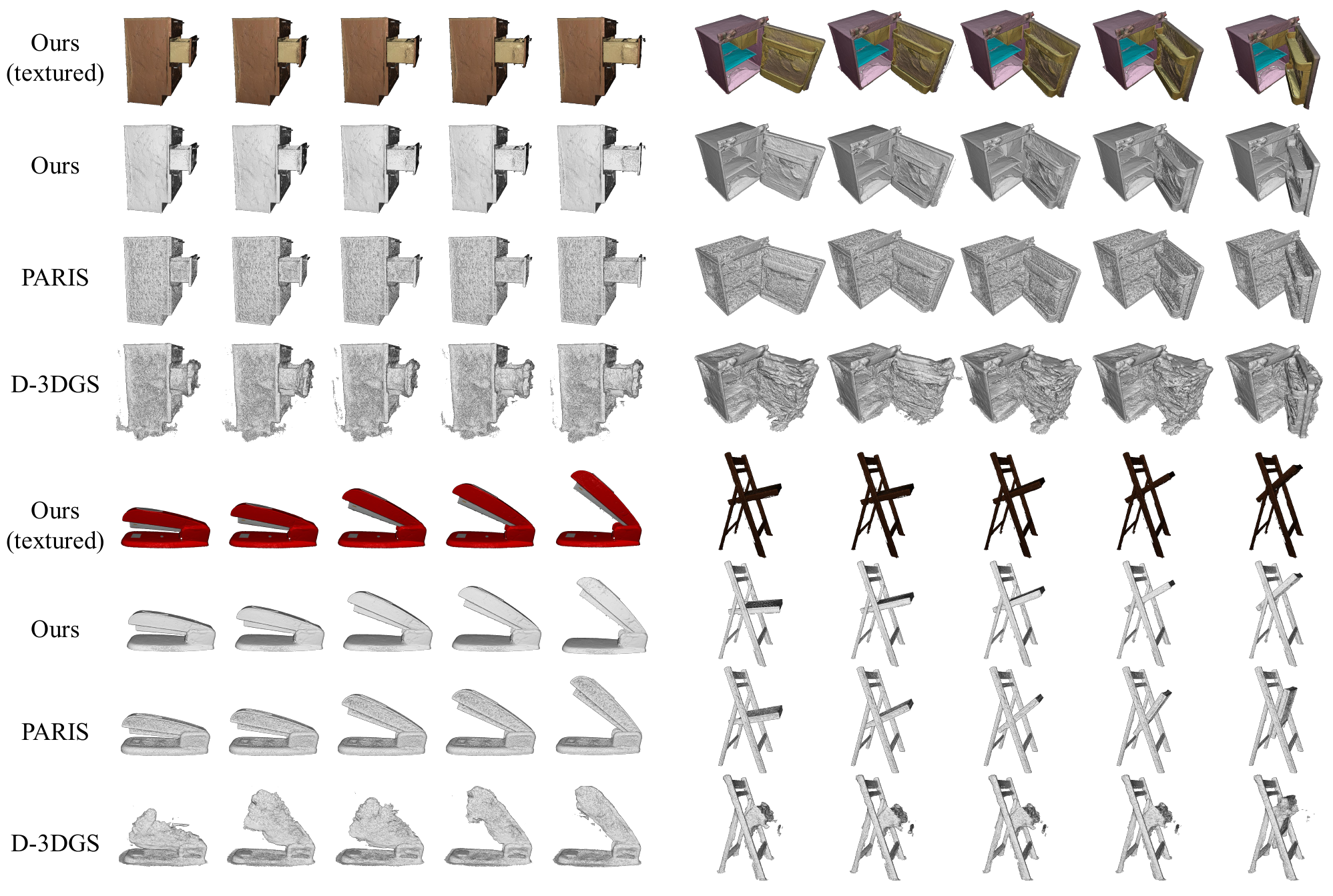}
\caption{The qualitative results of surface generation at arbitrary unseen states on PartNet-Mobility dataset. We show both textured and non-textured meshes for best comparison. The states are sampled randomly.}
\label{fig:unseen_generation}
\vspace{-1.0em}
\end{figure*} 

\begin{table*}[tbh]
\centering
\caption{Quantitative results for the surface generation quality on PartNet-Mobility dataset. We bold the best results and underline the second best results. $*$ means we implement it without depth supervision for fair comparison.}
\resizebox{0.95\linewidth}{!}{
\begin{tabular}{cc|ccccccccccc}
\toprule

Metrics   &Method      & Stapler  & USB &  Scissor & Fridge & Foldchair & Washer & Blade & Laptop & Oven & Storage & Mean \\
\hline

\multirow{5}{*}{CD(ws) $\downarrow$}
& A-SDF~\cite{mu2021sdf} &112.30 & 34.01& 94.11&33.46 & 56.73&55.82 &3.78 &27.64 &29.60 & 25.30&47.28 \\
& PARIS~\cite{liu2023paris} & \textbf{2.09}  &14.51 &12.81 &2.64 &10.42 & \underline{18.30} & 0.80& \underline{3.25} & 10.89 & \textbf{5.89} & 8.16 \\
& D-3DGS~\cite{yang2024deformable} &98.90 &16.79 & 91.84&48.78 &30.10 & 30.93&1.262 &65.11 &22.39 & 18.79& 42.49\\
& ArtGS$^{*}$~\cite{liu2025building} &\underline{2.34} &\underline{1.72} &\underline{0.68} &\textbf{2.05} &\underline{0.52}  &20.59 &\underline{0.63} &\textbf{1.00} &\underline{8.78} &10.97 &\underline{4.92} \\
& REArtGS (Ours)  &2.60 &\textbf{1.68} &\textbf{0.44} &\underline{2.21} &\textbf{0.41} & \textbf{14.80} & \textbf{0.59} & 3.65 & \textbf{8.11} & \underline{6.48} &\textbf{4.10}\\
\hline
\multirow{5}{*}{CD(rs) $\downarrow$} 
& A-SDF~\cite{mu2021sdf} &8.992 & 1.806& 1.822&5.166 & 5.668&7.050 & 5.991& 2.710&5.245 &7.794 &5.224 \\
& PARIS~\cite{liu2023paris}  &\textbf{2.113} &\textbf{0.417} &4.704 &\textbf{0.408} &1.216 &\textbf{0.857} &4.442 &1.354 &\underline{2.793} & 4.332 & 2.264\\
& D-3DGS~\cite{yang2024deformable} &9.589 &1.544 &0.891 &0.465 &\underline{1.147} &3.218 &\textbf{0.168} &\underline{0.449} &3.773 &3.100 &2.434 \\
& ArtGS$^{*}$~\cite{liu2025building} &4.924 &0.670 &\underline{0.381} &0.471 &1.622 &2.670 & 1.455 &1.269 &3.912 &\textbf{0.562} &\underline{1.794}\\
& REArtGS (Ours)  &\underline{3.651} &\underline{0.487} &\textbf{0.127} &\underline{0.464} &\textbf{0.420} &\underline{1.617} &\underline{0.414} &\textbf{0.298} & \textbf{2.129}& \underline{1.020}& \textbf{1.063}\\

\hline
\multirow{5}{*}{F1 $\uparrow$}
& A-SDF~\cite{mu2021sdf} &0.007 & 0.034& 0.056&0.012 &0.021 &0.023 & 0.168&0.001 &0.010 &0.004 & 0.034\\
& PARIS~\cite{liu2023paris} &0.221 &0.151 & 0.319 & 0.091 &0.423 &0.024 &0.421 &\underline{0.530} & 0.031& 0.032 & 0.224\\
& D-3DGS~\cite{yang2024deformable} &0.064 &0.180 &0.125 &0.073 &0.168 & \textbf{0.195}&0.233 &0.131 & 0.022&0.023 &0.121 \\
& ArtGS$^{*}$~\cite{liu2025building} &\textbf{0.264} &\textbf{0.255} &\underline{0.454} &\underline{0.125} &\underline{0.460} &\underline{0.055} &\underline{0.426} &0.292 &\underline{0.062} &\underline{0.045} &\underline{0.244} \\
& REArtGS (Ours) &\underline{0.254} &\underline{0.215} &\textbf{0.577} & \textbf{0.186} & \textbf{0.465} & 0.052 & \textbf{0.509}&\textbf{0.608} & \textbf{0.065}& \textbf{0.050} & \textbf{0.298}\\

\hline
\multirow{5}{*}{EMD $\downarrow$}  
& A-SDF~\cite{mu2021sdf} &\underline{2.123} &0.951 & 0.955&1.606 &1.684 &1.878 &1.731 &1.115 &1.620 &1.837 &1.559 \\
& PARIS~\cite{liu2023paris} &2.454 &\textbf{0.457} &1.533 &\textbf{0.451} &0.780 &\textbf{0.655} &1.491 &0.770 &\underline{1.183} &1.354 &1.113 \\
& D-3DGS~\cite{yang2024deformable} &3.224 &0.612 &0.665 &0.481 & 0.758&1.269 &\textbf{0.291} &\underline{0.436} &1.374 & 1.206& 1.032 \\
& ArtGS$^{*}$~\cite{liu2025building} &2.207 &0.562 &\underline{0.412} &0.563 &\underline{0.525} &1.171 &0.631 &0.724 &1.297 &\underline{0.750} &\underline{0.884}\\
& REArtGS (Ours)  & \textbf{1.594}&\underline{0.494} &\textbf{0.251} &\underline{0.473} &\textbf{0.459} &\underline{0.900} &\underline{0.456} &\textbf{0.353} &\textbf{1.033} &\textbf{0.689} &\textbf{0.670}\\

\bottomrule
\end{tabular}
}

\label{tab:exp_generation}
\vspace{-0.5cm}
\end{table*}

\section{Experiments}
\subsection{Experimental Setting}
\textbf{Datasets}. We conduct surface reconstruction experiments on synthetic dataset PartNet-Mobility~\cite{xiang2020sapien} and real-world dataset AKB-48~\cite{liu2022akb} to evaluate the reconstruction quality of our method. Please see the Appendix for more details.

\textbf{Metrics}.
Chamfer Distance (CD), F1-score and Earth Mover’s Distance (EMD)~\cite{zhang2022deepemd} are usually adopted as evaluation metrics for surface quality. However, these methods typically focus on the whole surface, endowing the unseen areas (such as the bottom of the object) with considerable weights during evaluation. This is unreasonable since we tend to be more concerned with the upper hemisphere regions of articulated objects. Therefore, we sample point clouds from the fragments between the rays cast by test cameras and the surface mesh to calculate EMD and CD, defined as CD (rs). We also evaluate the F1-score  and CD for the whole surface, defined as CD (ws), to comprehensively evaluate the surface quality. Note that the CD values are multiplied by 1000 and the distance threshold of F1-score is set to 0.4. The calculation of these metrics can be referred to in our Appendix.

All the experiments are conducted on a single RTX 4090 GPU. We also provide detailed implementation of our approach and baselines in the Appendix.

\vspace{-0.5em}
\subsection{Mesh Reconstruction Performance}

%We report the quantitative results of mesh reconstruction in Table \ref{tab:exp_reconstruction}. As can be observed, our REArtGS outperforms state-of-the-arts in four metrics by a big margin with only \textbf{3.79}, \textbf{1.236}, \textbf{0.294} and \textbf{0.695} respectively. Compared to A-SDF~\cite{mu2021sdf} that exploits implicit representation for articulated objects, REArtGS largely improves CD(ws) and CD(rs) metrics with \textbf{11.05} and \textbf{1.441}. Similar improvements also appear in comparisons with Ditto~\cite{jiang2022ditto} and PARIS~\cite{liu2023paris}. \textcolor{red}{add a script why REArtGS outperforms PARIS}. In terms of the 3DGS-based reconstruction method Deformable 3DGS, REArtGS also achieves better performance in the metrics and produce much more geometric details. Thus, we can conclude that within the geometric and motion constraints design, we can fully exploit 3DGS for the articulation mesh learning and analysis. From Table \ref{tab:exp_reconstruction}, we observe that the best reconstruction performance lies on \textit{Scissor}, \textit{Fridge} and \textit{Foldchair} categories, in which \textcolor{red}{add some descriptions and discussion.} Qualitative results are shown in Fig. \ref{fig:mesh_reconstruction}.
We report the quantitative results of mesh reconstruction in Table.~\ref{tab:exp_reconstruction} and qualitative results are shown in Fig.~\ref{fig:mesh_reconstruction}. Note that we implement ArtGS without depth supervision for a fair comparison. As can be observed, our REArtGS outperforms state-of-the-art approaches in the mean across all metrics with \textbf{3.79}, \textbf{1.236}, \textbf{0.294} and \textbf{0.695} respectively. Compared to A-SDF~\cite{mu2021sdf} and Ditto~\cite{jiang2022ditto} that exploits 3D inputs for articulated object surface reconstruction, our method still achieves higher reconstruction quality in the majority of categories. In terms of PARIS, GOF and ArtGS, our approach exhibits significantly smoother and clearer surfaces as shown in Fig.~\ref{fig:mesh_reconstruction}, benefiting from our enhanced geometric constraints. %As for the 3DGS-based reconstruction method GOF, our approach achieves marked superiority in the vast majority of categories. This is primarily attributed to our geometric constraints regularizing the opacity fields and facilitating more reasonable distribution of 3D Gaussian primitives over scene surfaces.

\subsection{Mesh Generation Performance}
We report the quantitative results of mesh generation in Table.~\ref{tab:exp_generation} and present the qualitative results Fig.~\ref{fig:unseen_generation}. Please refer to our Appendix for more qualitative results. Our REArtGS achieves the best mean results across all metrics with \textbf{4.10}, \textbf{1.063}, \textbf{0.298} and \textbf{0.670}. Compared to PARIS, our method exhibits smoother surface generation results, as shown in Fig.~\ref{fig:unseen_generation}. It is mainly because PARIS lacks a reasonable geometry initiation and directly utilizes the composite rendering, leading to the spatial overlap between the neural radiation fields of the start and end states. It can be also observed that our method is superior to ArtGS on most categories. This is primarily attributed to our high-quality geometry initialization from  the reconstruction stage, as well as the motion-constrained deformable fields to yield dynamic generation.

Besides, we provide the part segmentation and joint parameter estimation results in the Appendix and ~\ref{sec:joint_parameter} respectively.

\begin{minipage}[c]{0.5\textwidth}
\centering
\captionsetup{font=small, width=0.9\linewidth}
\captionof{table}{The ablation study of the unbiased SDF guidance on PartNet-Mobility dataset.}
 \label{tab:sdf_ablation}
\resizebox{0.9\linewidth}{!}{
\begin{tabular}{cc|cccc}
\toprule
w/ SDF  & w/ Unbiased Reg.  & CD (ws)  & CD (rs)   & F1   & EMD  \\
\hline
        &        & 5.96 & 2.242 & 0.250 & 1.366 \\                        
\checkmark &   & 4.38      & 1.921    & 0.273   & 0.817       \\
\checkmark & \checkmark & \textbf{3.79}     & \textbf{1.236}     & \textbf{0.294}   & \textbf{0.695} \\
\bottomrule
\end{tabular}}
\end{minipage}
\begin{minipage}[c]{0.5\textwidth}
\centering
\captionsetup{font=small, width=0.9\linewidth}
\captionof{table}{The ablation study of motion constraints on PartNet-Mobility dataset.}
\label{tab:motion_ablation} 
\resizebox{0.9\linewidth}{!}{\begin{tabular}{c|cccc}
\toprule
Settings  & CD (ws)  & CD (rs)   & F1   & EMD  \\
\hline
 w/o motion constraints   & 18.65  & 2.270 & 0.209   & 1.105     \\
 w/ motion constraints    & \textbf{5.41} & \textbf{1.063} & \textbf{0.276} & \textbf{0.670} \\  
\bottomrule
\end{tabular}}
\end{minipage}

%\vspace{-0.5em}
\subsection{Ablation Studies}
%鼓励高斯原语尽可能地贴近表面

\begin{wraptable}[18]{R}{0.55\textwidth}
%\begin{table}[tbh]
\centering
\vspace{-1.0em}
\caption{The quantitative results on real-world  AKB-48 dataset. We bold the best results and underline the second best results. $*$ means we implement it without depth supervision for fair comparison.}
\resizebox{1.0\linewidth}{!}{\begin{tabular}{cc|cccc|cccc}
\toprule
\multirow{2}{*}{Category} & \multirow{2}{*}{Method} & \multicolumn{4}{c|}{Reconstruction} & \multicolumn{4}{c}{Generation}\\
&& CD(ws)$\downarrow$ & CD(rs)$\downarrow$ & F1$\uparrow$ & EMD$\downarrow$ & CD(ws)$\downarrow$ & CD(rs)$\downarrow$ & F1$\uparrow$ & EMD$\downarrow$ \\
\midrule
\multirow{3}{*}{Box} & PARIS~\cite{liu2023paris} & \underline{4.69} &1.651 &0.097 &0.843 &\underline{3.98} &2.131 &0.306 &1.055 \\
& ArtGS$^{*}$~\cite{liu2025building}  &8.61 &\underline{0.967} &\underline{0.143} &\underline{0.697} &10.23 &\underline{1.606} &\underline{0.309} &\underline{0.851} \\
& REArtGS (Ours) &\textbf{2.49} & \textbf{0.898} &\textbf{0.205} &\textbf{0.671} &\textbf{1.28} &\textbf{1.425} &\textbf{0.559} &\textbf{0.845} \\   
\midrule
\multirow{3}{*}{Stapler} & PARIS~\cite{liu2023paris} &\textbf{0.23} &\underline{0.104} &\underline{0.551} &\underline{0.165} &\underline{0.47} &\underline{0.198} &\underline{0.540} &\underline{0.194} \\
& ArtGS$^{*}$~\cite{liu2025building} &51.04 &1.364 &0.476 &0.827 &5.13 &0.527 &0.533 &0.514\\
& REArtGS (Ours) &\underline{0.24} &\textbf{0.037} &\textbf{0.728} &\textbf{0.137} &\textbf{0.34} &\textbf{0.044} &\textbf{0.582} &\textbf{0.151} \\ 
\midrule
\multirow{3}{*}{Scissor} & PARIS~\cite{liu2023paris} &\underline{0.18} &0.302 &\underline{0.710} &0.314 &\underline{0.18} &0.329 &0.424 & 0.310\\
& ArtGS$^{*}$~\cite{liu2025building} &11.43 &\underline{0.073} &0.588 &\underline{0.192} &5.53 &\underline{0.089} &\underline{0.600} &\underline{0.213} \\
&REArtGS (Ours) &\textbf{0.08} &\textbf{0.017} &\textbf{0.902} &\textbf{0.093} &\textbf{0.08} &\textbf{0.003} &\textbf{0.899} &\textbf{0.042}
\\
\midrule
\multirow{3}{*}{Cutter} & PARIS~\cite{liu2023paris} & \underline{2.56} &1.371 &0.570 &\underline{0.302} &\textbf{45.02} &1.684 &\textbf{0.309} &1.183 \\
& ArtGS$^{*}$~\cite{liu2025building} &5.65 &\underline{0.453} &\underline{0.644} &0.477 &65.64 &\textbf{1.298} &0.093 &\textbf{0.806} \\
& REArtGS (Ours) &\textbf{0.13} &\textbf{0.016}  &\textbf{0.874} &\textbf{0.090} &\underline{51.56} &\underline{1.559} &\underline{0.228} &\underline{0.884}\\
\midrule
\multirow{3}{*}{Drawer} & PARIS~\cite{liu2023paris} &30.63 &\underline{4.568} &0.062 &\underline{1.577} &33.94 &3.865 &0.091 &1.430 \\
& ArtGS$^{*}$~\cite{liu2025building} &\underline{19.41} &7.478 &\underline{0.086} &1.935 &\underline{12.60} &\textbf{1.624} &\underline{0.094} &\textbf{0.911}\\
& REArtGS (Ours) &\textbf{10.73} &\textbf{2.607} &\textbf{0.138} &\textbf{1.142} &\textbf{11.71} &\underline{1.739} &\textbf{0.209} &\underline{0.933}\\
\midrule
\multirow{3}{*}{Eyeglasses} & PARIS~\cite{liu2023paris} &\underline{0.16} &1.582 &\underline{0.637} &0.212 &\underline{0.18} &\textbf{1.550} &0.609 &1.284 \\
& ArtGS$^{*}$~\cite{liu2025building} &27.49 &\underline{1.384} &0.557 &\underline{0.832} &12.12 &1.592 &\underline{0.747} &\underline{0.955} \\
& REArtGS (Ours) &\textbf{0.04} &\textbf{0.018} &\textbf{0.969} &\textbf{0.096} &\textbf{0.04} &\underline{1.568} &\textbf{0.974} &\textbf{0.886}\\
\midrule
\multirow{3}{*}{Mean} & PARIS~\cite{liu2023paris} &\underline{6.41} &\underline{1.336} &\underline{0.438} &\underline{0.569} & \underline{13.96}&1.626 &0.380 &\underline{0.707}\\
& ArtGS$^{*}$~\cite{liu2025building} &20.61 &1.953 &0.416 &0.826 &18.54 &\underline{1.123} &\underline{0.396} &0.708 \\
& REArtGS (Ours) &\textbf{2.86} &\textbf{0.599} &\textbf{0.636} &\textbf{0.372} &\textbf{10.83} &\textbf{1.056} &\textbf{0.575} &\textbf{0.624}\\
\bottomrule
\end{tabular}
}

\label{tab:exp_realworld}
%\vspace{-0.5cm}
%\end{table}
\end{wraptable}

We conduct an ablation of the unbiased SDF guidance on PartNet-Mobility dataset to validate its effectiveness. The ablation setting and results are reported in Table.~\ref{tab:sdf_ablation}. The quantitative results prove that both SDF guidance and the unbiased regularization of SDF indeed improve surface reconstruction quality. This is primarily because the SDF representation establishes explicit geometric association between the opacity fields and scene surface through $\Phi_{k}$, and the unbiased SDF regularization further optimizes the distribution of Gaussian opacity fields.

% SDF guidence
% \begin{table}[tbh]
% \centering
% \caption{The ablation study of the unbiased SDF guidance on PartNet-Mobility dataset.}
% \resizebox{0.6\linewidth}{!}{\begin{tabular}{cc|cccc}
% \toprule
% w/ SDF  & w/ Unbiased Reg.  & CD (ws)  & CD (rs)   & F1   & EMD  \\
% \hline
%         &        & 5.96 & 2.242 & 0.250 & 1.366 \\                        
% \checkmark &   & 4.38      & 1.921    & 0.273   & 0.817       \\
% \checkmark & \checkmark & \textbf{3.79}     & \textbf{1.236}     & \textbf{0.294}   & \textbf{0.695} \\
% \bottomrule
% \end{tabular}}

% \label{tab:sdf_ablation} 
% %\vspace{-0.3cm}
% \end{table}

% \begin{wraptable}[4]{r}{0.6\textwidth}
% \centering
% \caption{The ablation study of the unbiased SDF guidance on PartNet-Mobility dataset.}
% \begin{tabular}{cc|cccc}
% \toprule
% w/ SDF  & w/ Unbiased Reg.  & CD (ws)  & CD (rs)   & F1   & EMD  \\
% \hline
%         &        & 5.96 & 2.242 & 0.250 & 1.366 \\                        
% \checkmark &   & 4.38      & 1.921    & 0.273   & 0.817       \\
% \checkmark & \checkmark & \textbf{3.79}     & \textbf{1.236}     & \textbf{0.294}   & \textbf{0.695} \\
% \bottomrule
% \end{tabular}
% \label{tab:sdf_ablation}
% \end{wraptable}

% D-3DGS motion constraints
We also conduct an ablation study of the motion constraints on PartNet-Mobility dataset and present the quantitative results in Table.~\ref{tab:motion_ablation}. For the baseline without motion constraints, we employ three MLPs to learn the variations of spatial positions, scales, and quaternions respectively for Gaussian primitives, which is similar to D-3DGS. The significantly superior results demonstrate the effectiveness of the motion constraints. %Actually, motion constraints effectively compress the parameter search space of the deformation fields according to the kinematic structures of articulated objects and accurately infer the deformed spatial positions via rigid transformation.

% \begin{table}[tbh]
% \centering
% \caption{The ablation study of motion constraints. We compare the surface generation quality on the PartNet-Mobility dataset between training with motion constraints and without motion constraints.}
% \resizebox{0.6\linewidth}{!}{\begin{tabular}{c|cccc}
% \toprule
% Settings  & CD (ws)  & CD (rs)   & F1   & EMD  \\
% \hline
%  w/o motion constraints   & 18.65  & 2.270 & 0.209   & 1.105     \\
%  w/ motion constraints    & \textbf{5.41} & \textbf{1.063} & \textbf{0.276} & \textbf{0.670} \\  
% \bottomrule
% \end{tabular}}

% \label{tab:motion_ablation} 
% \vspace{-0.3cm}
% \end{table}

\vspace{-0.2cm}
\begin{figure}[tbh]
    \centering
    %\vspace{-1.2em}
    \includegraphics[width=0.9\linewidth]{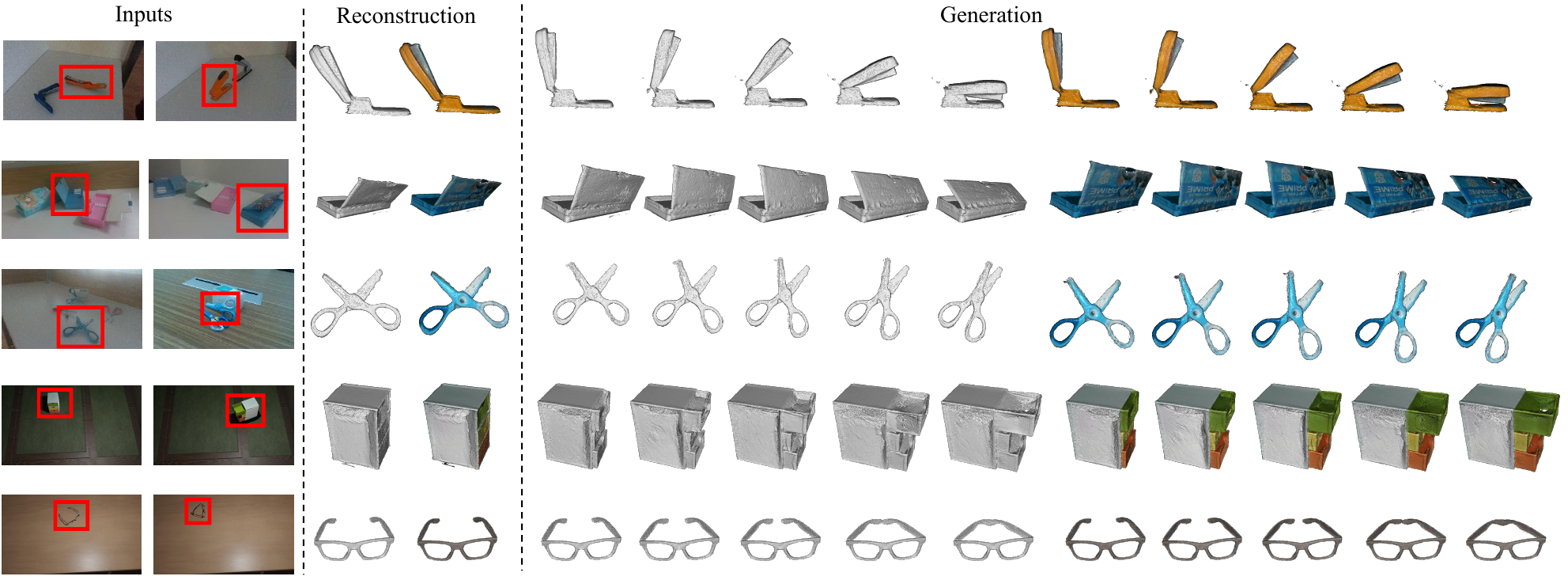}
    %\vspace{-0.5em}

    \caption{The qualitative results of surface reconstruction on real-world AKB-48~\cite{liu2022akb} repository. We show both textured and non-textured meshes for two-part and multi-part articulated objects.} %The multi-view images are obtained from ReArtVal~\cite{liu2022toward} dataset.
       \label{fig:real_world}
       \vspace{-0.53cm}
\end{figure} 

\subsection{Generalization to the Real World}

We conduct reconstruction and generation experiments in the real world to investigate the generalization capacity of our REArtGS. We report the quantitative results comparing with PARIS and ArtGS in Table.~\ref{tab:exp_realworld}, and present the qualitative results in Fig.~\ref{fig:real_world}. Our approach significantly outperforms PARIS and ArtGS in mean performance across all metrics. This demonstrates that our method exhibits strong generalization capability for real-world objects and enables high-quality surface reconstruction and generation as shown in Fig.~\ref{fig:real_world}.

\section{Conclusion and Future Work}
\label{sec:conlusion}
We propose REArtGS, a novel framework that achieves high-quality textured mesh reconstruction and dynamic generation of articulated objects using only RGB views of two arbitrary states. We propose an unbiased SDF guidance to regularize Gaussian opacity fields and model the Gaussian deformation fields constrained by kinematic structures for generating meshes of unseen states. Experiments show the superior performance of our method on both synthetic and real-world data. 

The limitations of REArtGS lie on the requirement of camera pose prior and challenges of objects with transparent materials. Future works will introduce relative pose estimation to alleviate the dependence on camera poses and employ physically-based networks to model the transparent materials.

\section*{Acknowledgements}
This work was supported by National Natural Science Foundation of China under Grant 62302143, 62576130, and Anhui Province's Key Science and Technology Project 202423k09020007.

\bibliographystyle{plain}
\bibliography{main}

\end{document}